\documentclass{article} 
\usepackage{iclr2027_conference,times}

\usepackage{amsmath,amsfonts,bm}

\def\eqref#1{equation~\ref{#1}}

\def\1{\bm{1}}

\DeclareMathAlphabet{\mathsfit}{\encodingdefault}{\sfdefault}{m}{sl}
\SetMathAlphabet{\mathsfit}{bold}{\encodingdefault}{\sfdefault}{bx}{n}

\usepackage{hyperref}
\usepackage{enumitem}
\usepackage{url}
\usepackage{amsmath, amsfonts, amssymb}
\usepackage{amsthm}
\newtheorem{theorem}{Theorem}

\usepackage{enumitem}
\usepackage{booktabs}
\usepackage{multirow} 
\usepackage{xcolor}
\usepackage{colortbl}
\usepackage[cal=cm]{mathalpha}
\usepackage{mathrsfs}
\usepackage{graphicx}
\usepackage{subcaption}
\definecolor{c1}{HTML}{d497c3}   
\definecolor{c2}{HTML}{FAC7C8}  
\definecolor{c3}{HTML}{FDEEEE} 
\providecommand{\cI}{\cellcolor{c1}}
\providecommand{\cII}{\cellcolor{c2}}
\providecommand{\cIII}{\cellcolor{c3}}

\title{Trajectory-Aware Policy Optimization:\\ Closing the State Distribution Gap in Long-Horizon LLM RL}

\author{Zhicheng Cai, Hanlin Wu, Weinan Dai, Yuyan Zhou, Wei-Ying Ma, Ya-Qin Zhang, Hao Zhou\thanks{Corresponding author: \texttt{zhouhao@air.tsinghua.edu.cn}}\\
Institute for AI Industry Research (AIR), Tsinghua University\\
SIA-Lab of Tsinghua AIR and ByteDance Seed
}

\iclrfinalcopy 
\begin{document}

\maketitle

\begin{abstract}
Reinforcement learning (RL) has become a dominant paradigm for enhancing LLMs' reasoning capabilities.
Yet long-horizon autoregressive rollouts introduce a fundamental limitation: local policy mismatch can compound into severe state distribution shift over trajectories.
Existing LLM RL methods, such as GRPO and GSPO, control policy mismatch at the token or sequence level, but overlook this trajectory-induced shift, causing training instability and eventual collapse as the horizon grows.
To close the state distribution gap, we propose \emph{Trajectory-Aware Policy Optimization (TAPO)}, a principled framework that theoretically derives the state-distribution correction from the off-policy policy gradient and incorporates it into the RL objective.
TAPO further employs a geometric-mean estimator for variance reduction, together with a length-aware dynamic clipping that prevents overly deviated updates across the trajectory.
Extensive experiments demonstrate that TAPO significantly surpasses existing LLM RL algorithms across eight competition-level benchmarks, achieving up to 48\% and 12\% improvement over GRPO on mathematical reasoning and algorithmic coding, with substantially enhanced training stability.
\end{abstract}

\section{Introduction}
Reinforcement Learning (RL) has emerged as a critical paradigm for eliciting complex reasoning capabilities in Large Language Models (LLMs)~\citep{guo2025deepseek,comanici2025gemini,openai2024learningtoreason}, yielding major breakthroughs across long-horizon reasoning frontiers including mathematical problem-solving~\citep{shao2024deepseekmath,wang2024math,luo2023wizardmath}, algorithmic coding~\citep{dou2024stepcoder,anthropic2024claudecode}, and deep search~\citep{openai2025deepresearch,jin2025searchr1}.

\begin{figure}[t]
  \centering
  \includegraphics[width=1.0\textwidth]{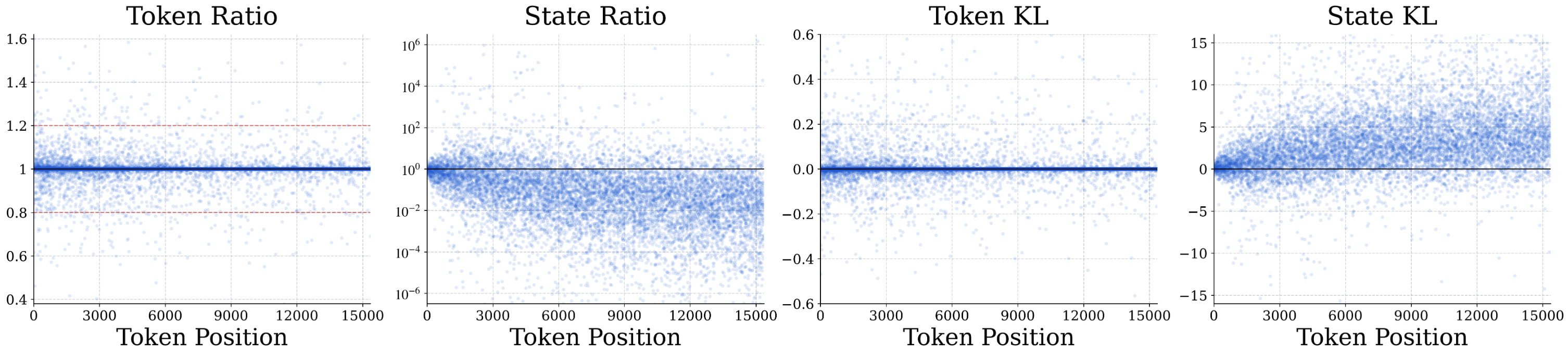}
  \vspace{-1em}
  \caption{Policy shifts along long-horizon token positions, measured by token- and state-level policy ratios and KL divergences. The dots are randomly sampled through the training process of GRPO.
  While the token-level ratio remains relatively stable across positions, the state-level ratio progressively deviates from unity toward later positions, revealing the accumulation of policy shift over long-horizon trajectories.}
  \label{fig:shift}
\end{figure}

However, long-horizon autoregressive rollouts expose RL to a fundamental vulnerability: local policy mismatch can compound into severe state distribution shift over the trajectory.
At each step, the target policy acts on a state induced by all previously sampled tokens, making its state distribution dependent on the entire history of policy deviations.
Thus, even modest deviations at individual positions can propagate through the trajectory, progressively driving the target policy toward states increasingly distant from the behavior policy's state distribution. 
As empirically demonstrated in Fig.~\ref{fig:shift}, while the local token-level ratio remains relatively controllable across positions, the state-level ratio progressively drifts away from unity over long horizons.

\begin{figure}[t]
  \centering
  \includegraphics[width=\textwidth]{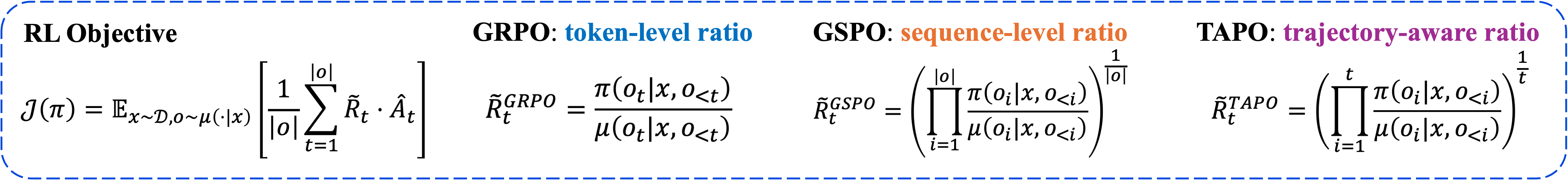}
  \includegraphics[width=\textwidth]{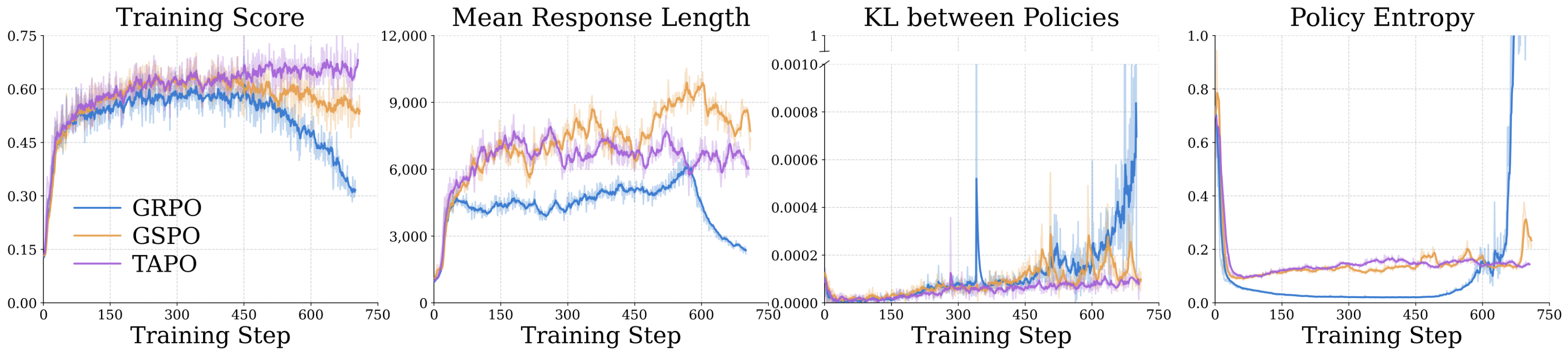}
  \vspace{-1em}
  \caption{Formulations (upper) and training dynamics (bottom) of GRPO, GSPO, and TAPO. GRPO and GSPO apply policy-ratio correction at the token and sequence levels, respectively, whereas TAPO explicitly incorporates the state-distribution correction induced by the preceding trajectory. As response lengths increase, GRPO and GSPO exhibit growing KL divergence and eventual training collapse, while TAPO maintains controlled KL divergence and stable improvement.}
  \label{fig:main}
\end{figure}

Importance sampling provides a principled mechanism for correcting off-policy sampling mismatch, yet existing LLM RL methods primarily perform this correction at the action level. 
Two representative approaches, GRPO~\citep{shao2024deepseekmath} and GSPO~\citep{zheng2025group}, operate at different granularities: GRPO applies token-level importance ratios, while GSPO regards the full sequence as an action and aggregates token-level ratios into a sequence-level correction. 
However, neither explicitly corrects for the state distribution shift induced by the preceding trajectory. 
Consequently, as response lengths grow during training, the accumulated state shift manifests as an increasing KL divergence between the behavior and target policies, eventually driving training into collapse, as shown in Fig.~\ref{fig:main}. 
This exposes a fundamental gap in existing policy optimization: controlling action-level mismatch at the token or sequence level does not guarantee control of the state distribution shift induced over long-horizon trajectories.

To close this gap, we propose \textit{Trajectory-Aware Policy Optimization} (TAPO), a theoretically grounded RL algorithm that fundamentally corrects the state distribution shift over the trajectory.
We first derive the off-policy policy gradient to reveal the state-distribution correction required for long-horizon autoregressive trajectories, and incorporate this correction into the RL objective. 
To control the variance of the state-level correction, we use a geometric-mean estimator over the preceding trajectory, yielding a stable and length-robust trajectory-aware policy ratio (Fig~\ref{fig:main}).
Building on this formulation, we further introduce a dynamic clipping mechanism that adapts the clipping boundary to the length of preceding trajectory, preventing overly aggressive updates as the horizon grows while preserving sufficient flexibility at early explorations.

We conduct extensive experiments across various LLMs and competition-level benchmarks for mathematical reasoning and algorithmic coding.
TAPO consistently outperforms existing RL algorithms, achieving up to 48\% and 12\% average improvements over GRPO on math and code, respectively, demonstrating substantial gains in long-horizon reasoning performance.
Moreover, as Fig.~\ref{fig:main} shows, while GRPO and GSPO eventually suffer from training collapse as the policy shift grows, TAPO maintains a steadily increasing training score with controlled KL divergence.
These results demonstrate that explicitly correcting the trajectory-induced state distribution shift not only improves reasoning performance but also substantially enhances training stability.

In summary, our contributions are threefold:
\vspace{-0.5em}
\begin{itemize}[leftmargin=*]
\item We identify and formalize the accumulation of state distribution shift as a fundamental issue in long-horizon LLM RL, and derive its corresponding correction from the off-policy policy gradient.
\item We propose TAPO, a theoretically grounded RL algorithm, employing geometric-mean trajectory-aware ratio together with length-aware dynamic clipping for stable long-horizon optimization.
\item Extensive experiments across mathematical reasoning and algorithmic coding demonstrate that TAPO consistently improves reasoning performance with enhanced training stability.
\end{itemize}

\section{Preliminary}
\subsection{MDP Formulation for Autoregressive LLMs}
The autoregressive generation of an LLM can be formulated as a finite-horizon Markov decision process (MDP).
Given a prompt $x$, at step $t$, the state $s_t=(x,o_{<t})$ is defined by prompt together with the preceding generated tokens $o_{<t}=(o_1,\ldots,o_{t-1})$, and the action $o_t\sim\pi(\cdot\mid s_t)$ is the $t$-th token sampled from the policy $\pi$ (parameterized by $\theta$).
A trajectory is therefore given by $\tau=(s_1,o_1,\ldots,o_T,s_{T+1})\sim P^\pi(\cdot|x)$, with the next state deterministically induced by the current action as $s_{t+1}=(x,o_{\leq t})$. Thus, the terminal state is $s_{T+1}=(x,o)$, where $o=(o_1,\ldots,o_T)$ denotes the complete response.

Typically, LLM reasoning tasks provide only an outcome reward \(R(\tau)\) for the entire response trajectory.
The objective of RL is to maximize the expected reward over generated trajectories:
\begin{equation}
\mathcal{J}(\pi) = \mathbb{E}_{x\sim\mathcal{D},\,\tau\sim P^\pi(\cdot|x)} \left[ R(\tau) \right],\ \quad \nabla_\theta \mathcal{J}(\pi)=\mathbb{E}_{x\sim\mathcal{D},\,\tau\sim P^\pi(\cdot|x)}\left[\sum_{t=1}^{T}\nabla_\theta \log \pi(o_t\mid s_t){A}_t\right]
\end{equation}
where the gradient is obtained by the policy gradient theorem~\citep{NIPS1999_464d828b}, and \(A_t\) denotes the advantage assigned to the token $o_t$.

In practice, LLM RL methods typically sample the trajectories from the behavior policy $\mu$ while optimize the target policy $\pi$, yielding an off-policy mismatch that requires correction.

\subsection{GRPO: Token-Level Correction}
Group Relative Policy Optimization (GRPO)~\citep{shao2024deepseekmath,guo2025deepseek} follows PPO~\citep{schulman2017proximal} in applying an action-level importance ratio $r_t^g$ to individual tokens to correct the off-policy mismatch.
Specifically, the GRPO objective is given by:
\begin{equation}
\begin{aligned}
\mathcal{J}_\text{GRPO}(\pi)=\mathbb{E}_{x\sim\mathcal{D},\{o^g\}^G_{g=1}\sim\mu(\cdot|x)}&\Bigg[\frac{1}{G}\sum^G_{g=1}\frac{1}{|o^g|}\sum^{|o^g|}_{t=1}\min\left(r_t^g\hat{A}_t^g, \text{clip}\left(r_t^g,1\!-\!\epsilon^l,1\!+\!\epsilon^h\right)\hat{A}_t^g\right)\Bigg]\\
&\text{where}\quad r_t^g=\frac{\pi(o_t^g\mid x,o_{<t}^g)}{\mu(o_t^g\mid x,o_{<t}^g)}
\label{eq:grpo}
\end{aligned}
\end{equation}
GRPO estimates advantages in a group-relative manner, where the behavior policy $\mu$ samples a group of $G$ individual responses $\{o^g\}_{g=1}^G$ for a given query, and the advantage assigned to the $g$-th response is obtained by normalizing its reward $R^g$ within the group, thus $\hat{A}_t^g = \frac{R^g - \text{mean}(\{R^g\}_{g=1}^G)}{\text{std}(\{R^g\}_{g=1}^G)}$.
Moreover, subsequent GRPO variants, such as DAPO~\citep{yu2025dapo}, DCPO~\citep{yang2025dcpo}, DPPO~\citep{qi2026rethinking}, and RIPO~\citep{cai2026beyond} maintain the token-level importance ratio.

\subsection{GSPO: Sequence-Level Correction}
Group Sequence Policy Optimization (GSPO)~\citep{zheng2025group} instead regards the entire response as a single action to match the unit of sequence-level reward, and applies importance-ratio correction at the sequence level.
Specifically, GSPO employs the following optimization objective:
\begin{equation}
\begin{aligned}
\mathcal{J}_\text{GSPO}(\pi)&=\mathbb{E}_{x\sim\mathcal{D},\{o^g\}^G_{g=1}\sim\mu(\cdot|x)}\Bigg[\frac{1}{G}\sum^G_{g=1}\min\left(s^{g}\hat{A}^{g}, \text{clip}\left(s^{g},1\!-\!\epsilon^l,1\!+\!\epsilon^h\right)\hat{A}^{g}\right)\Bigg]\\
&=\mathbb{E}_{x\sim\mathcal{D},\{o^g\}^G_{g=1}\sim\mu(\cdot|x)}\Bigg[\frac{1}{G}\sum^G_{g=1}\frac{1}{|o^g|}\sum^{|o^g|}_{t=1}\min\left(s_t^g\hat{A}_t^g, \text{clip}\left(s_t^g,1\!-\!\epsilon^l,1\!+\!\epsilon^h\right)\hat{A}_t^g\right)\Bigg]\\
&\text{where}\quad s_t^g=\text{sg}[s^g]\frac{\pi(o_t^g\mid x,o_{<t}^g)}{\text{sg}[\pi(o_t^g\mid x,o_{<t}^g)]},\quad s^g=\left(\prod^{|o^g|}_{i=1}\frac{\pi(o_i^g\mid x,o_{<i}^g)}{\mu(o_i^g\mid x,o_{<i}^g)}\right)^{\frac{1}{|o^g|}}
\label{eq:gspo}
\end{aligned}
\end{equation}
$sg[\cdot]$ denotes only taking the numerical value but stopping the gradient, $s^g$ is the sequence-level likelihood of importance ratio, and the term $\frac{\pi(o_t^g\mid x,o_{<t}^g)}{\text{sg}[\pi(o_t^g\mid x,o_{<t}^g)]}$ with identity value is used to assign the gradient to the token at position $t$.

However, GSPO does not fully align with the autoregressive MDP structure of LLM RL, where each token is generated as an action at a state induced by the preceding trajectory. 
Treating the entire trajectory as the unit of clipping also leads to low token efficiency: once the sequence-level ratio falls outside the clipping range, all tokens in the trajectory are clipped, including intermediate tokens whose policy mismatch may remain well controlled.
Conversely, even when the sequence-level ratio remains within the clipping range, substantial state distribution shifts may still occur at intermediate positions. Although these local deviations can be averaged out in the final sequence-level ratio, the corresponding token-level updates are still influenced by these severely shifted states, potentially causing unstable optimization.

Moreover, despite their different correction granularities, both GRPO and GSPO correct policy mismatch through action likelihoods, without explicitly accounting for the state distribution induced by the preceding trajectory.
We next show that a principled off-policy policy gradient necessarily involves an additional state-distribution correction induced by the preceding trajectory.

\section{Method}
In this section, we propose Trajectory-Aware Policy Optimization (TAPO) to correct the state distribution shift induced by preceding trajectory.
We first derive the off-policy policy gradient to establish the theoretical importance ratio, then develop a variance-reduced correction and a length-aware dynamic clipping for stable training, and finally integrate these components into the unified TAPO objective.

\subsection{From Action Correction to State Correction}
For a target policy $\pi$ optimized using trajectories sampled from a behavior policy $\mu$, the objective can be equivalently evaluated under the behavior policy via trajectory-level importance sampling:
\begin{equation}
\mathcal{J}(\pi) = \mathbb{E}_{x\sim\mathcal{D},\,\tau\sim P^\pi(\cdot|x)} \left[ R(\tau) \right]=\mathbb{E}_{x\sim\mathcal{D},\,\tau\sim P^\mu(\cdot|x)} \left[\left(\frac{P^\pi(\tau|x)}{P^\mu(\tau|x)}\right) R(\tau) \right]
\end{equation}
The resulting off-policy policy gradient is given by the following theorem.~\footnote{The detailed derivation is deferred to Appendix.}
\begin{theorem}[Off-Policy Policy Gradient]
Given target policy $\pi$ to be optimized and behavior policy $\mu$ for sampling trajectories, the off-policy policy gradient is
\begin{equation}
\nabla_\theta\mathcal{J}(\pi)=\mathbb{E}_{x\sim\mathcal{D}}\left[\sum_{t=1}^{T}\mathbb{E}_{s_t\sim d_t^\mu(\cdot|x),\,o_t\sim\mu(\cdot|s_t)}\left[\frac{d_t^\pi(s_t|x)}{d_t^\mu(s_t|x)}\frac{\pi(o_t|s_t)}{\mu(o_t|s_t)}\nabla_\theta\log\pi(o_t|s_t)A^\pi(s_t,o_t)\right]\right]
\end{equation}
where $A^\pi(s_t,o_t)$ is the advantage function, $d_t^\pi(s_t|x)\triangleq\Pr_\pi(s_t|x)=\sum_{\tau_{<t}:s_t}P^\pi(\tau_{<t}|x)$ and $d_t^\mu(s_t|x)\triangleq\Pr_\mu(s_t|x)=\sum_{\tau_{<t}:s_t}P^\mu(\tau_{<t}|x)$ denote the state distributions induced by $\pi$ and $\mu$ at step $t$, respectively.
\end{theorem}

The theorem reveals that the exact off-policy correction consists of two components: a state-distribution correction $\frac{d_t^\pi(s_t|x)}{d_t^\mu(s_t|x)}$ and an action-level correction $\frac{\pi(o_t|s_t)}{\mu(o_t|s_t)}$. 
In traditional RL, policy optimization is typically formulated in continuing or infinite-horizon MDPs, where the state distributions are characterized by a single MDP-induced stationary distribution~\citep{schulman2015high,schulman2017proximal}. 
Under this formulation, state-distribution mismatch is handled implicitly through the policy-induced stationary distribution rather than explicitly corrected at each time step, and practical off-policy methods therefore often focus on action-level importance ratios.

In contrast, autoregressive LLM generation forms a finite-horizon, tree-structured MDP, where each generated token extends the state and thereby determines the branch of all subsequent states. 
There is no single stationary state distribution that characterizes the states encountered at different token positions. 
Consequently, local policy mismatch at preceding positions accumulates along the trajectory and can induce substantial state-distribution mismatch at later positions, making the state-distribution correction non-negligible for long-horizon autoregressive generation.

Specifically, the state transition of LLM RL is deterministic, each state $s_t=(x,o_{<t})$ uniquely identifies the preceding token sequence, thus the state-distribution ratio of LLM reduces to the likelihood ratio of the preceding trajectory:
\begin{equation}
\frac{d_t^\pi(s_t|x)}{d_t^\mu(s_t|x)}A=\frac{\sum_{\tau_{<t}:s_t}P^\pi(\tau_{<t}|x)}{\sum_{\tau_{<t}:s_t}P^\mu(\tau_{<t}|x)}=\frac{\prod_{i=1}^{t-1}\pi(o_i|s_i)}{\prod_{i=1}^{t-1}\mu(o_i|s_i)}=\prod_{i=1}^{t-1}\frac{\pi(o_i|s_i)}{\mu(o_i|s_i)}.
\end{equation}
Consequently, the theoretically required importance correction $\mathcal{R}_t$ at token $t$ can be written as
\begin{equation}
\mathcal{R}_t\triangleq\underbrace{\frac{d_t^\pi(s_t|x)}{d_t^\mu(s_t|x)}}_{\text{state correction}}
\cdot\underbrace{\frac{\pi(o_t\mid s_t)}{\mu(o_t\mid s_t)}}_{\text{action correction}}=\prod_{i=1}^{t}\frac{\pi(o_i\mid s_i)}{\mu(o_i\mid s_i)}.
\end{equation}

\subsection{Trajectory-Aware Policy Optimization}
Based on the theoretically required importance correction derived above, we first formulate a trajectory-aware corrected surrogate objective for LLM RL:
\begin{equation}
\begin{aligned}
\mathcal{J}_{traj}(\pi)=\mathbb{E}_{x\sim\mathcal{D},\{o^g\}^G_{g=1}\sim\mu(\cdot|x)}\Bigg[\frac{1}{G}\sum^G_{g=1}\frac{1}{|o^g|}\sum^{|o^g|}_{t=1}\min\left(\mathcal{R}_t^g\hat{A}_t^g, \text{clip}\left(\mathcal{R}_t^g,1\!-\!\epsilon^l,1\!+\!\epsilon^h\right)\hat{A}_t^g\right)\Bigg]
\label{eq:traj}
\end{aligned}
\end{equation}
While the ratio $\mathcal{R}_t^g=\prod_{i=1}^{t}\frac{\pi(o_i^g\mid s_i^g)}{\mu(o_i^g\mid s_i^g)}$ provides the theoretically required trajectory-aware correction, directly using the product of per-token likelihood ratios can lead to high variance and unstable optimization, especially for long-horizon generation. 
Moreover, both the variance and the effective range of $\mathcal{R}_t^g$ vary with the length of the preceding trajectory, making the fixed clipping boundaries used by GRPO and GSPO inadequate. We therefore introduce two complementary mechanisms to make trajectory-aware correction practical: variance-controlled trajectory correction and length-aware dynamic clipping.
\subsubsection{Variance-Controlled Trajectory Correction}
The correction ratio $\mathcal{R}_t$ is a product of $t$ per-token likelihood ratios, whose variance can grow rapidly with trajectory length, thus making the training unstable. 
To control this variance while preserving the trajectory-level correction structure, we replace the product with its geometric mean (we omit the superscript $g$ for clarity):
\begin{equation}
\widetilde{\mathcal{R}}_t \triangleq \left( \prod_{i=1}^{t} \frac{\pi(o_i\mid s_i)}{\mu(o_i\mid s_i)} \right)^{1/t} = \mathcal{R}_t^{1/t}.
\end{equation}
The variance reduction induced by geometric averaging can be characterized as follows.
\begin{theorem}[Variance of Trajectory Correction Ratio]
Let $r_i=\frac{\pi(o_i\mid s_i)}{\mu(o_i\mid s_i)}$ and suppose $\log r_i$ are i.i.d. Gaussian with variance $\sigma^2$. Since $\mathbb{E}_{\mu}[r_i]=1$, we have $\log r_i\sim\mathcal{N}(-\frac{\sigma^2}{2},\sigma^2)$. Then the exact trajectory-level correction and its geometric-mean counterpart satisfy
\begin{equation}
\operatorname{Var}_{\mu}(\mathcal{R}_t) = e^{t\sigma^2}-1, \qquad \operatorname{Var}_{\mu}(\widetilde{\mathcal{R}}_t) = e^{-\sigma^2+\sigma^2/t}\left(e^{\sigma^2/t}-1\right).
\end{equation}
Hence, the variance of the product correction grows exponentially with the length of preceding trajectory, whereas the variance of the geometric-mean correction decays as $O(1/t)$.
\end{theorem}

\subsubsection{Length-Aware Dynamic Clipping}
Geometric averaging resolves the variance explosion of the exact trajectory correction, but it also makes the correction increasingly concentrated as the trajectory grows.
In particular, the distribution of $\widetilde{\mathcal{R}}_t$ becomes progressively narrower in log-space, implying that a fixed clipping boundary cannot provide a consistent level of tolerated proximal region across different token positions.
We therefore derive a length-aware clipping rule from this concentration property.

\begin{theorem}[Concentration of Geometric-Mean Correction]
\label{theorem:t2}
Let $X_i=\log r_i$ be independent random variables such that $X_i-\mathbb{E}[X_i]$ is $\sigma^2$-sub-Gaussian, i.e.,
\begin{equation}
\mathbb{E}\left[\exp\left(\lambda(X_i-\mathbb{E}[X_i])\right)\right]\leq\exp\left(\frac{\lambda^2\sigma^2}{2}\right),\qquad \forall\lambda\in\mathbb{R}.
\end{equation}
For the geometric-mean correction $\widetilde{\mathcal{R}}_t=\exp\left(\frac{1}{t}\sum_{i=1}^{t}X_i\right)$, its log-ratio satisfies
\begin{equation}
\Pr\left(\left|\log\widetilde{\mathcal{R}}_t-\mathbb{E}[\log\widetilde{\mathcal{R}}_t]\right|\geq\epsilon\right)\leq2\exp\left(-\frac{t\epsilon^2}{2\sigma^2}\right).
\end{equation}
\end{theorem}

Thus, for a desired coverage probability $1-\delta$, the deviation radius in log-space scales as $\epsilon_t=\sigma\sqrt{\frac{2\log(2/\delta)}{t}}=O\left(\frac{1}{\sqrt{t}}\right)$.
Therefore, the effective range of $\widetilde{\mathcal{R}}_t$ contracts with trajectory length: early tokens require a wider tolerance, whereas later tokens admit a tighter clipping range.
This perspective is consistent with the clipping boundaries adopted by existing methods: GRPO uses clipping boundaries on the order of $10^{-1}$ for token-level ratios, whereas GSPO uses substantially tighter boundaries on the order of $10^{-4}$ for sequence-level ratios.

Theorem~\ref{theorem:t2} motivates us to adopt a length-aware clipping boundary whose length-dependent scaling follows $1/\sqrt{t}$.
Specifically,
\begin{equation}
\epsilon_t^{l}=\epsilon_{\min}^{l}+(\epsilon_{\max}^{l}-\epsilon_{\min}^{l})\frac{1/\sqrt{t}-1/\sqrt{T}}{1-1/\sqrt{T}},\qquad \epsilon_t^{h}=\epsilon_{\min}^{h}+(\epsilon_{\max}^{h}-\epsilon_{\min}^{h})\frac{1/\sqrt{t}-1/\sqrt{T}}{1-1/\sqrt{T}}.
\label{epsilon}
\end{equation}
The endpoint normalization maps the $1/\sqrt{t}$ scaling to the interval $[1,T]$, ensuring that the clipping range starts from a relatively permissive boundary $\epsilon_{\max}$ and gradually tightens toward a conservative boundary $\epsilon_{\min}$ as the trajectory grows.

\subsubsection{Trajectory-Aware Policy Optimization}
Combining variance-controlled trajectory correction with length-aware dynamic clipping, we obtain the final TAPO objective:
\begin{equation}
\begin{aligned}
\mathcal{J}_\text{TAPO}(\pi) &=\mathbb{E}_{x\sim\mathcal{D},\{o^g\}_{g=1}^G\sim\mu(\cdot|x)}\Bigg[\frac{1}{G}\sum_{g=1}^G\frac{1}{|o^g|}\sum_{t=1}^{|o^g|}\min\left(\widehat{\mathcal{R}}_t^g\hat{A}_t^g,\operatorname{clip}\left(\widehat{\mathcal{R}}_t^g,1-\epsilon_t^l,1+\epsilon_t^h\right)\hat{A}_t^g\right)\Bigg]\\
&\text{where}\quad \widehat{\mathcal{R}}_t^g=\operatorname{sg}\left[\widetilde{\mathcal{R}}_t^g\right]\frac{\pi(o_t^g\mid x,o_{<t}^g)}{\operatorname{sg}\left[\pi(o_t^g\mid x,o_{<t}^g)\right]},\quad \widetilde{\mathcal{R}}_t^g=\left(\prod_{i=1}^{t}\frac{\pi(o_i^g\mid x,o_{<i}^g)}{\mu(o_i^g\mid x,o_{<i}^g)}\right)^{1/t}.
\end{aligned}
\label{eq:TAPO}
\end{equation}
where $\operatorname{sg}[\cdot]$ denotes the stop-gradient operator, $\epsilon_t^l$ and $\epsilon_t^h$ follow the length-aware schedule defined in Eq.~\ref{epsilon}.
Although $\widehat{\mathcal{R}}_t^g$ has the same forward value as $\widetilde{\mathcal{R}}_t^g$, its gradient satisfies
\begin{equation}
\nabla_\theta\widehat{\mathcal{R}}_t^g=\widetilde{\mathcal{R}}_t^g\nabla_\theta\log\pi(o_t^g\mid x,o_{<t}^g),
\end{equation}
which decouples the trajectory-level correction from the gradient of preceding actions and retains the local policy-gradient structure of the current action.
This construction is consistent with the off-policy policy gradient derived in Theorem~1: the trajectory-level ratio corrects the accumulated distribution shift, while the current action likelihood provides the policy-gradient direction.

By comparison, GRPO is action-corrected but state-agnostic, using only the local likelihood ratio for each token, whereas GSPO applies a single sequence-level correction uniformly across all tokens.
TAPO instead provides trajectory-aware, token-specific correction, weighting each token according to the accumulated policy shift along the trajectory leading to its current state. 
Consequently, TAPO offers a principled middle ground between local action correction and global sequence correction, explicitly bridging the state-distribution gap in long-horizon autoregressive generation and enabling consistently stable training, as demonstrated in Fig.~\ref{fig:main}.

\begin{table}[!t]
\caption{Comparison of different RL algorithms trained on Qwen3-8B-Base across various mathematical reasoning and algorithmic coding benchmarks (evaluated by Avg@8). We highlight each cell as \colorbox{c1}{best}, \colorbox{c2}{second best}, and \colorbox{c3}{third best}.}
\vspace{-1em}
\label{tab:main}
\begin{center}
\small
\begin{tabular}{ll|cccc|c}
\toprule
\multicolumn{7}{c}{\textbf{Mathematical Reasoning}} \\
\toprule
Method & Granularity & AIME24 & AIME25 & BRUMO25 & SMT25 & Average \\
\midrule
Base   & none & 3.8   & 3.3   & 15.0  & 10.8  & \ \ 8.2 (-70.4\%) \\
GRPO   &  token-level  & 34.6  & 20.8  & 33.3  & 21.9  & 27.7 (\ \ +0.0\%) \\
DAPO   &  token-level  & 45.0  & 27.5  & 37.5  & 28.5  & 34.6 (+24.9\%) \\
DCPO   &  token-level  & 36.3  & 27.5  & \cIII 42.8  & 31.6  & 34.6 (+24.9\%) \\
SAPO   &  token-level  & 37.3  & 28.7  & 41.7  & 32.3  & 35.0 (+26.4\%) \\
CISPO  &  token-level  & 31.7  & 23.8  & 38.3  & 28.0  & 30.5 (+10.1\%) \\
Clip-Cov  &  token-level  & 36.3  & 22.9  & 36.6  & 30.2  & 31.5 (+13.7\%) \\
DPPO   &  token-level  & \cIII 46.7  & \cII 32.5  & 42.5  & \cI 34.2  & \cIII 39.0 (+40.8\%) \\
RIPO   &  token-level  & \cII 47.9  & \cII 32.5  & \cI 47.5  & \cI 34.2  & \cII 40.5 (+46.2\%) \\
GSPO   &  sequence-level  & 42.9  & 28.7  & 36.3  & 32.5  & 35.1 (+26.7\%) \\
GMPO   &  sequence-level  & 41.7  & \cIII 30.0  & 35.8  & \cIII 33.0  & 35.1 (+26.7\%) \\
\textbf{TAPO}  &  trajectory-aware  & \cI 49.6  & \cI 33.3  & \cII 47.2 & \cII 33.8 & \cI 41.0 (+48.0\%) \\
\toprule
\multicolumn{7}{c}{\textbf{Code Generation}} \\
\toprule
Method & Granularity  & Codeforces & CodeContests & TACO & APPS & Average \\
\midrule
Base   & none  & 32.3  & 32.1  & 16.5  & 32.1  & 28.3 (-45.9\%) \\
GRPO   & token-level & 61.9  & 57.4  & 36.2  & 54.2  & 52.4 (\ \ +0.0\%) \\
DAPO   & token-level & \cIII 64.0  & 59.1  & 37.4  & \cIII 56.8  & 54.3 (\ \ +3.6\%) \\
DCPO   & token-level & 63.5  & 58.7  & 36.8  & \cII 57.2  & 54.1 (\ \ +3.2\%) \\
SAPO   & token-level & 63.7  & 57.0  & 37.5  & 55.9  & 53.5 (\ \ +2.1\%) \\
CISPO  & token-level & 61.6  & 54.6  & 35.1  & 53.9  & 51.3 (\ \ -2.1\%) \\
Clip-Cov  & token-level & 62.9 & 57.4 & 36.0 & 55.5  & 53.0 (\ \ +1.2\%) \\
DPPO   & token-level & 63.5  & 58.6  & \cIII 37.6  & 55.6  & 53.8 (\ \ +2.7\%) \\
RIPO   & token-level & \cII 68.2  & \cII 62.3  & \cII 39.3  & \cI 58.9  & \cII 57.2 (\ \ +9.2\%) \\
GSPO   & sequence-level & 63.8  & 57.9  & 36.6  & 56.3  & 53.7 (\ \ +2.5\%) \\
GMPO   & sequence-level & \cIII 64.0  & \cIII 59.2  & 37.4  & \cIII 56.8  & \cIII 54.4 (\ \ +3.8\%) \\
\textbf{TAPO}  & trajectory-aware & \cI 70.2 & \cI 64.5 & \cI 40.7  & \cI 58.9  & \cI 58.6 (+11.8\%) \\
\bottomrule
\end{tabular}
\end{center}
\vspace{-1em}
\end{table}
\section{Experiments}
\vspace{-0.5em}
To validate the effectiveness of TAPO, we conduct extensive experiments across eight competition-level benchmarks spanning both mathematical reasoning and algorithmic coding, with comparisons against ten representative RL algorithms on models of different scales.
\vspace{-0.5em}
\subsection{Experimental Setup}
\vspace{-0.5em}
\textbf{Baselines.}
We compare the proposed TAPO to ten representative RL algorithms spanning different correction granularities of policy shift. For token-level correction, we compare against GRPO~\citep{shao2024deepseekmath}, DAPO~\citep{yu2025dapo}, DCPO~\citep{yang2025dcpo}, SAPO~\citep{gao2025soft}, CISPO~\citep{chen2025minimax}, Clip-Cov~\citep{cui2025entropy}, DPPO~\citep{qi2026rethinking}, and RIPO~\citep{cai2026beyond}
; for sequence-level correction, we compare against GSPO~\citep{zheng2025group} and GMPO~\citep{zhao2025geometric}.
All compared methods follow their default hyperparameter settings.

Following common practice~\citep{ye2020mastering,cai2026beyond}, we apply dual clipping to the importance ratio with lower and upper bounds of $0.5$ and $10$, and remove the KL penalty term for all methods~\citep{yu2025dapo,cai2026beyond}.
All algorithms adopt the GRPO-style group-relative advantage estimator, and trained on Qwen3-8B-Base~\citep{yang2025qwen3}.
For TAPO, we set $\epsilon^l_{\max}\!=\!\epsilon^h_{\max}=0.2$ and $\epsilon^l_{\min}\!=\!0.0003, \epsilon^h_{\min}=0.0004$, respectively.

\textbf{Training.}
We evaluate the improvement induced by each RL algorithm on two complementary domains, mathematical reasoning and algorithmic coding, both of which admit verifiable rewards and are widely regarded as strong indicators of general reasoning ability.
For mathematical reasoning we train on DAPO-Math-17k~\citep{yu2025dapo}, and for algorithmic coding on Eurus-Code-25k~\citep{cui2025process}.
For each query we generate 8 rollouts (namely, the group size is 8) with a maximum response length of 20k tokens for thorough reasoning.
In each RL iteration, the behavior policy $\mu$ totally generates 1024 rollouts with a train batch size of 128, and the current policy $\pi_\theta$ is updated 8 times with a mini-batch size of 16.
Rewards are set to $1$ for correct responses and $0$ otherwise.
We uniformly adopt token-mean loss aggregation and optimize all models to convergence using AdamW~\citep{loshchilov2017decoupled} with a constant learning rate of $1\!\times\!10^{-6}$, training for 750 steps.
All experiments are conducted on VeRL~\citep{sheng2025hybridflow} framework using 8$\times$H20 GPUs.

\textbf{Evaluation.}
We comprehensively evaluate the effectiveness of these algorithms on eight competition-level benchmarks without contamination, spanning the two domains.
For math we use AIME24, AIME25, BRUMO25, and SMT25~\citep{balunovic2025matharena}; for coding we use Codeforces, CodeContests, TACO, and APPS~\citep{cui2025process}.
We evaluate the best checkpoint of each algorithm, repeat the evaluation set for 8 times and report $\text{avg}@8$ metric for results stability. 
\vspace{-0.5em}
\subsection{Main Results}
\vspace{-0.5em}
Table~\ref{tab:main} summarizes the performance of different RL methods across eight competition-level mathematical reasoning and algorithmic coding benchmarks.
TAPO consistently achieves the best performance across both task domains. Specifically, compared to GRPO, TAPO improves the average performance by $48.0\%$ on math and $11.8\%$ on coding. Compared to the strongest token-level correction method, RIPO, TAPO further improves the average performance by $1.2\%$ and $2.4\%$, respectively. Moerover, compared to the sequence-level method GSPO, TAPO achieves improvements of $16.8\%$ and $9.1\%$ on math and coding.

Moreover, Fig.~\ref{fig:main} shows the training dynamics of GRPO, GSPO, and TAPO on mathematical reasoning. As observed, GRPO and GSPO eventually suffer from training collapse, characterized by declining training scores and fluctuating entropy as the policy shift grows ( reflected by the increasing KL divergence). In contrast, TAPO maintains steadily improving training scores with controlled KL divergence and stable entropy throughout training, validating that trajectory-aware correction effectively mitigates the state distribution shift induced by long-horizon policy updates, leading to more stable and sustained training.

Overall, the experimental results persuasively demonstrate the effectiveness and generality of trajectory-aware policy optimization.
\vspace{-0.5em}
\subsection{Generalization Across Model Scales}
\vspace{-0.5em}
\begin{table}[t]
\caption{Comparison of different RL algorithms trained on Qwen3-1.7B and Qwen3-4B across various mathematical reasoning benchmarks (evaluated by Avg@8). We highlight each cell as \colorbox{c1}{best}, \colorbox{c2}{second best}, and \colorbox{c3}{third best}.}
\label{tab:math}
\begin{center}
\small
\vspace{-1em}
\begin{tabular}{ll|cccc|c}
\toprule
Method & Granularity & AIME24 & AIME25 & BRUMO25 & SMT25 & Average\\
\midrule
\multicolumn{7}{c}{\textit{Qwen3-1.7B-Base}}\\
\midrule
Base & none & 2.1  & 0.0  & 11.3 & 6.6  & \ \ 5.0 (-55.0\%)  \\
GRPO & token-level & 11.3 & 10.8 & \cIII 15.0 & 7.3  & 11.1 (\ \ +0.0\%)\\
DAPO & token-level & 15.0 & \cIII 12.1 & 9.2  & 7.8  & 11.0 (\ \ -0.7\%)\\
DCPO & token-level & 17.1 & 11.3 & \cII 15.4 & \cI 10.8 & \cIII 13.7 (+23.0\%)\\
RIPO & token-level & \cII 18.3 & \cI 12.9 & \cII 15.4 & \cII 10.4 & \cII 14.3 (+28.4\%)\\
GSPO & sequence-level & 16.3 & 11.2 & 12.9 & \cIII 8.5  & 12.2 (+10.1\%) \\
GMPO & sequence-level & \cIII 17.5 & \cII 12.5 & \cIII 15.0 & 6.6  & 12.9 (+16.2\%) \\
\textbf{TAPO}  & trajectory-aware & \cI 18.9 & \cII 12.5 & \cI 16.2 & \cI 10.8 & \cI 14.6 (+31.5\%) \\
\midrule
\multicolumn{7}{c}{\textit{Qwen3-4B-Base}}\\
\midrule
Base & none & 10.0 & 8.3  & 18.3 & 8.9  & 11.4 (-53.0\%)\\
GRPO & token-level & 30.4 & 20.8 & 25.8 & 19.8 & 24.2 (\ \ +0.0\%)\\
DAPO & token-level & 31.7 & 21.3 & 26.3 & \cII 21.7 & 25.3 (\ \ +4.3\%)\\
DCPO & token-level & 31.7 & 24.2 & 25.0 & 19.8 & 25.2 (\ \ +4.0\%) \\
RIPO & token-level & \cI 32.9 & \cII 27.5 & \cII 31.3 & \cIII 21.5 & \cII 28.3 (+16.9\%) \\
GSPO & sequence-level & \cIII 32.1 & 21.3 & \cIII 26.7 & 18.2 & 24.6 (\ \ +1.5\%)\\
GMPO & sequence-level & \cII 32.5 & \cIII 26.7 & 26.3 & \cI 22.5 & \cIII 27.0 (+11.6\%) \\
\textbf{TAPO}  & trajectory-aware & \cI 32.9 & \cI 28.7 & \cI 33.3 & \cI 22.5 & \cI 29.5 (+21.3\%) \\
\bottomrule
\end{tabular}
\vspace{-1em}
\end{center}
\end{table}
We further generalize TAPO to different model scales, including Qwen3-1.7B-Base and Qwen3-4B-Base, trained on mathematical reasoning.
As shown in Table~\ref{tab:math}, TAPO consistently achieves the best average performance on both models, improving over GRPO by $31.5\%$ and $21.3\%$, respectively. 
Compared with GMPO, TAPO further improves the average performance by $13.2\%$ on Qwen3-1.7B and $9.3\%$ on Qwen3-4B.
These results demonstrate that the general effectiveness of trajectory-aware correction across model scales.
\vspace{-0.5em}
\subsection{Ablation Studies}
\vspace{-0.5em}
\begin{figure}[t]
  \centering
  \includegraphics[width=0.99\textwidth]{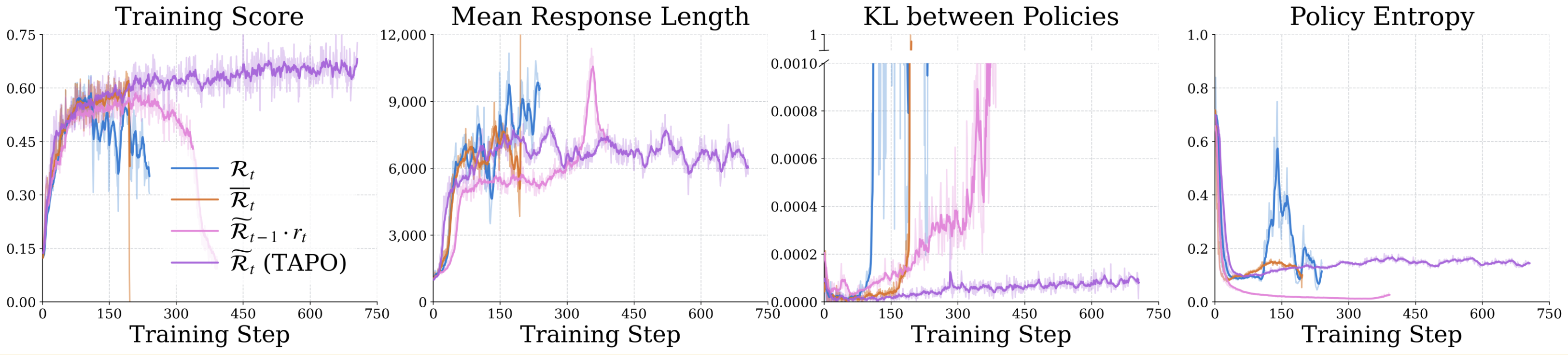}
  \includegraphics[width=0.99\textwidth]{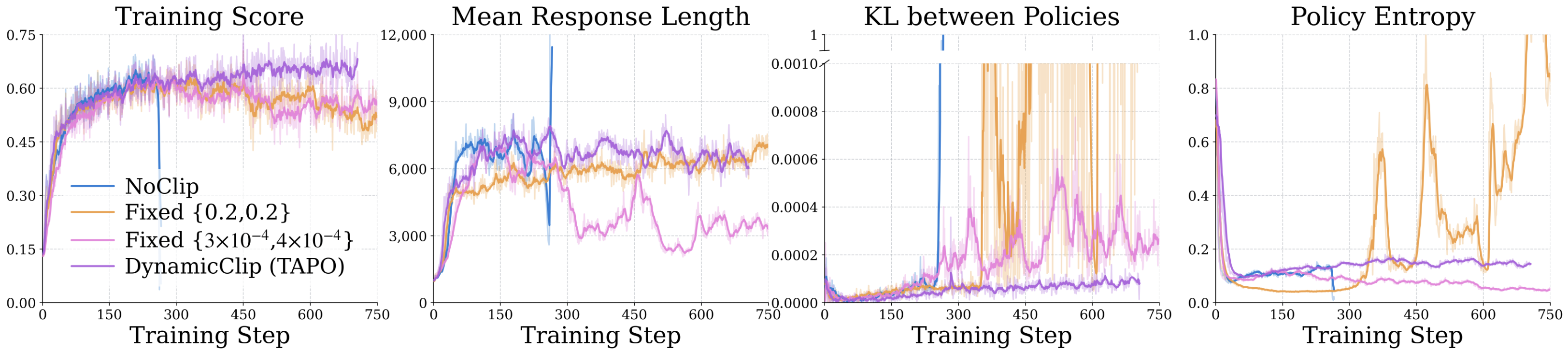}
  \vspace{-0.5em}
  \caption{Training dynamics of different ablation studies on mathematical reasoning tasks. \textbf{Upper}: trajectory correction ratio variants. \textbf{Lower}: clipping strategies.}
  \label{fig:as}
  \vspace{-1em}
\end{figure}
We further conduct ablation studies on Qwen3-8B-Base trained for mathematical reasoning to investigate the design choices of TAPO. Specifically, we examine two components: the construction of the trajectory-aware correction ratio and the length-aware clipping strategy.

\textbf{Trajectory-aware Correction Ratio.  }
For the trajectory correction ratio, we compare four variants: the original trajectory ratio $\mathcal{R}_t$; $\mathcal{R}_t$ with excessive samples outside the range $[e^{-0.4},e^{0.4}]$ removed, following GMPO (denoted as $\overline{\mathcal{R}}_t$); geometric averaging applied only to the state correction ratio, resulting in $\widetilde{\mathcal{R}}_{t-1}\cdot r_t$; and geometric averaging applied to the full trajectory ratio $\mathcal{R}_t$, corresponding to the proposed $\widetilde{\mathcal{R}}_t$.
As shown in the upper panel of Fig.~\ref{fig:as}, the original trajectory ratio $\mathcal{R}_t$ exhibits highly unstable training dynamics due to the unbounded variance introduced by multiplying correction ratios over the trajectory. Removing excessively large correction ratios following GMPO does not resolve this issue and eventually still leads to training collapse. Applying geometric averaging only to the state correction ratio, i.e., $\widetilde{\mathcal{R}}_{t-1}\cdot r_t$, essentially add an additional scaling factor to the action-level correction, making the clipping range difficult to calibrate and ultimately ineffective. In contrast, geometric averaging applied to the full trajectory ratio yields stable training dynamics and sustained score improvement. These results demonstrate that controlling the variance of the complete trajectory correction is essential for effective trajectory-aware policy optimization.

\textbf{Clipping Startegy.  }
We further compare four clipping strategies: no clipping; fixed clipping with $\epsilon^l_t\!=\!\epsilon^h_t\!=\!0.2$ (as adopted by GRPO); fixed clipping with $\epsilon^l_t\!=\!0.0003, \epsilon^h_t\!=\!0.0004$ (as adopted by GSPO); and the proposed length-aware dynamic clipping.
As shown in the lower panel of Fig.~\ref{fig:as}, removing the proximal constraint of clipping leads to sudden training collapse. 
GRPO-order clipping boundary provides insufficient constraint for later tokens, resulting in increasingly unstable KL divergence and entropy. In contrast, GSPO-order clipping boundary imposes overly restrictive constraints on earlier tokens, causing entropy to decrease rapidly and slowing score improvement. 
These results demonstrate the importance of adapting the clipping boundary to trajectory length.

\vspace{-1em}
\section{Conclusion}
\vspace{-0.5em}
This paper introduces Trajectory-Aware Policy Optimization (TAPO), a principled framework for correcting the policy distribution shift in long-horizon LLM RL, derived from the off-policy policy gradient.
Across eight mathematical reasoning and algorithmic coding benchmarks, TAPO substantially outperforms existing RL algorithms, demonstrating the effectiveness of trajectory-aware policy correction.
Ultimately, TAPO provides a principled foundation for advancing long-horizon autoregressive policy optimization.


\clearpage



\bibliography{iclr2027_conference}
\bibliographystyle{iclr2027_conference}

\clearpage
\appendix
\section{Proof of Theorem1}
Given the objective and a behavior policy $\mu$ for sampling, the trajectory distribution induced by policy $\mu$ can be written as $P^\mu(\tau|x)=\prod^T_{t=1}\mu(o_t|s_t)p(s_{t+1}|s_t,o_t)$. For LLM RL, the state transition is deterministic, and thus $p(s_{t+1}|s_t,o_t)=1$, yielding $P^\mu(\tau|x)=\prod^T_{t=1}\mu(o_t|s_t)$. Similarly, $P^\pi(\tau|x)=\prod^T_{t=1}\pi(o_t|s_t)$ denotes the trajectory distribution induced by the target policy $\pi$.
The objective under the target policy is
\begin{equation}
\mathcal{J}(\pi)=\mathbb{E}_{x\sim\mathcal{D},\,\tau\sim P^\pi(\cdot|x)}\left[R(\tau)\right].
\end{equation}
By importance sampling, this expectation can be evaluated using trajectories sampled from the behavior policy:
\begin{equation}
\begin{aligned}
\mathcal{J}(\pi)&=\mathbb{E}_{x\sim\mathcal{D},\,\tau\sim P^\mu(\cdot|x)}\left[\frac{P^\pi(\tau|x)}{P^\mu(\tau|x)}R(\tau)\right]\\
&=\mathbb{E}_{x\sim\mathcal{D},\,\tau\sim P^\mu(\cdot|x)}\left[\left(\prod_{t=1}^{T}\frac{\pi(o_t|s_t)}{\mu(o_t|s_t)}\right)R(\tau)\right].
\end{aligned}
\label{eq:app_is_objective}
\end{equation}
Taking the gradient with respect to the policy parameters $\theta$ gives
\begin{equation}
\nabla_\theta\mathcal{J}(\pi)=\mathbb{E}_{x\sim\mathcal{D},\,\tau\sim P^\mu(\cdot|x)}\left[\nabla_\theta\left(\prod_{t=1}^{T}\frac{\pi(o_t|s_t)}{\mu(o_t|s_t)}\right)R(\tau)\right].
\label{eq:app_gradient_product}
\end{equation}
Since the behavior policy $\mu$ is fixed during optimization, only the target policy probabilities depend on $\theta$. Applying the product rule yields
\begin{equation}
\nabla_\theta\left(\prod_{i=1}^{T}\frac{\pi(o_i|s_i)}{\mu(o_i|s_i)}\right)=\sum_{t=1}^{T}\left(\prod_{i\neq t}\frac{\pi(o_i|s_i)}{\mu(o_i|s_i)}\right)\frac{\nabla_\theta\pi(o_t|s_t)}{\mu(o_t|s_t)}.
\label{eq:app_product_rule}
\end{equation}
Using the log-derivative identity
\begin{equation}
\nabla_\theta\pi(o_t|s_t)=\pi(o_t|s_t)\nabla_\theta\log\pi(o_t|s_t),
\label{eq:app_log_derivative}
\end{equation}
we obtain
\begin{equation}
\begin{aligned}
\nabla_\theta\left(\prod_{i=1}^{T}\frac{\pi(o_i|s_i)}{\mu(o_i|s_i)}\right)&=\sum_{t=1}^{T}\left(\prod_{i\neq t}\frac{\pi(o_i|s_i)}{\mu(o_i|s_i)}\right)\frac{\pi(o_t|s_t)}{\mu(o_t|s_t)}\nabla_\theta\log\pi(o_t|s_t)\\
&=\left(\prod_{i=1}^{T}\frac{\pi(o_i|s_i)}{\mu(o_i|s_i)}\right)\sum_{t=1}^{T}\nabla_\theta\log\pi(o_t|s_t).
\end{aligned}
\label{eq:app_log_derivative_product}
\end{equation}
Substituting Eq.~\eqref{eq:app_log_derivative_product} into Eq.~\eqref{eq:app_gradient_product} gives
\begin{equation}
\nabla_\theta\mathcal{J}(\pi)=\mathbb{E}_{x\sim\mathcal{D},\,\tau\sim P^\mu(\cdot|x)}\left[\left(\prod_{i=1}^{T}\frac{\pi(o_i|s_i)}{\mu(o_i|s_i)}\right)\sum_{t=1}^{T}\nabla_\theta\log\pi(o_t|s_t)R(\tau)\right].
\label{eq:app_full_gradient}
\end{equation}
By linearity of expectation, this can be written as
\begin{equation}
\nabla_\theta\mathcal{J}(\pi)=\sum_{t=1}^{T}\mathbb{E}_{x\sim\mathcal{D},\,\tau\sim P^\mu(\cdot|x)}\left[\left(\prod_{i=1}^{T}\frac{\pi(o_i|s_i)}{\mu(o_i|s_i)}\right)\nabla_\theta\log\pi(o_t|s_t)R(\tau)\right].
\label{eq:app_gradient_sum}
\end{equation}

To derive a step-wise expression, consider the contribution of a fixed time step $t$. We decompose the trajectory as $\tau=(\tau_{<t},s_t,o_t,\tau_{>t})$. The full importance ratio can then be factorized as
\begin{equation}
\prod_{i=1}^{T}\frac{\pi(o_i|s_i)}{\mu(o_i|s_i)}=\left(\prod_{i=1}^{t}\frac{\pi(o_i|s_i)}{\mu(o_i|s_i)}\right)\left(\prod_{i=t+1}^{T}\frac{\pi(o_i|s_i)}{\mu(o_i|s_i)}\right).
\label{eq:app_ratio_factorization}
\end{equation}
Conditioning on $(s_t,o_t)$, the first factor and $\nabla_\theta\log\pi(o_t|s_t)$ are fixed. Therefore, the contribution at step $t$ becomes
\begin{equation}
\begin{aligned}
&\mathbb{E}_{\tau\sim P^\mu}\left[\left(\prod_{i=1}^{T}\frac{\pi(o_i|s_i)}{\mu(o_i|s_i)}\right)\nabla_\theta\log\pi(o_t|s_t)R(\tau)\right]\\
&=\mathbb{E}_{\tau\sim P^\mu}\left[\left(\prod_{i=1}^{t}\frac{\pi(o_i|s_i)}{\mu(o_i|s_i)}\right)\nabla_\theta\log\pi(o_t|s_t)\left(\prod_{i=t+1}^{T}\frac{\pi(o_i|s_i)}{\mu(o_i|s_i)}\right)R(\tau)\right].
\end{aligned}
\label{eq:app_step_contribution}
\end{equation}

Taking the conditional expectation over the future trajectory $\tau_{>t}$ gives
\begin{equation}
\begin{aligned}
&\mathbb{E}_{\tau_{>t}\sim P^\mu(\cdot|s_t,o_t)}\left[\left.\left(\prod_{i=t+1}^{T}\frac{\pi(o_i|s_i)}{\mu(o_i|s_i)}\right)R(\tau)\right|s_t,o_t\right]\\
&=\mathbb{E}_{\tau_{>t}\sim P^\pi(\cdot|s_t,o_t)}\left[R(\tau)\mid s_t,o_t\right]\\
&=Q^\pi(s_t,o_t).
\end{aligned}
\label{eq:app_future_expectation}
\end{equation}
The first equality follows from importance sampling applied to the future trajectory distribution, while the second equality follows from the definition of the action-value function.

Substituting Eq.~\eqref{eq:app_future_expectation} into Eq.~\eqref{eq:app_step_contribution} and summing over all time steps yields
\begin{equation}
\nabla_\theta\mathcal{J}(\pi)=\mathbb{E}_{x\sim\mathcal{D},\,\tau\sim P^\mu(\cdot|x)}\left[\sum_{t=1}^{T}\left(\prod_{i=1}^{t}\frac{\pi(o_i|s_i)}{\mu(o_i|s_i)}\right)\nabla_\theta\log\pi(o_t|s_t)Q^\pi(s_t,o_t)\right].
\label{eq:app_q_gradient}
\end{equation}

We next introduce the state-dependent value function
\begin{equation}
V^\pi(s_t)=\mathbb{E}_{o_t\sim\pi(\cdot|s_t)}\left[Q^\pi(s_t,o_t)\right],
\end{equation}
and define the advantage function as
\begin{equation}
A^\pi(s_t,o_t)=Q^\pi(s_t,o_t)-V^\pi(s_t).
\end{equation}
For each time step, subtracting $V^\pi(s_t)$ as a baseline does not change the gradient. Specifically, conditioned on $s_t$, we have
\begin{equation}
\begin{aligned}
&\mathbb{E}_{o_t\sim\pi(\cdot|s_t)}\left[\nabla_\theta\log\pi(o_t|s_t)V^\pi(s_t)\right]\\
&=V^\pi(s_t)\sum_{o_t}\pi(o_t|s_t)\nabla_\theta\log\pi(o_t|s_t)\\
&=V^\pi(s_t)\sum_{o_t}\nabla_\theta\pi(o_t|s_t)\\
&=V^\pi(s_t)\nabla_\theta\sum_{o_t}\pi(o_t|s_t)\\
&=0.
\end{aligned}
\label{eq:app_baseline}
\end{equation}
Thus, Eq.~\eqref{eq:app_q_gradient} can equivalently be expressed using the advantage function:
\begin{equation}
\nabla_\theta\mathcal{J}(\pi)=\mathbb{E}_{x\sim\mathcal{D},\,\tau\sim P^\mu(\cdot|x)}\left[\sum_{t=1}^{T}\left(\prod_{i=1}^{t}\frac{\pi(o_i|s_i)}{\mu(o_i|s_i)}\right)\nabla_\theta\log\pi(o_t|s_t)A^\pi(s_t,o_t)\right].
\label{eq:app_advantage_gradient}
\end{equation}

Finally, we marginalize the expectation at each time step over the state and action distributions induced by the behavior policy. For any function $f(s_t,o_t)$, we have
\begin{equation}
\mathbb{E}_{\tau\sim P^\mu(\cdot|x)}\left[\left(\prod_{i=1}^{t}\frac{\pi(o_i|s_i)}{\mu(o_i|s_i)}\right)f(s_t,o_t)\right]=\mathbb{E}_{s_t\sim d_t^\mu(\cdot|x),\,o_t\sim\mu(\cdot|s_t)}\left[\left(\prod_{i=1}^{t}\frac{\pi(o_i|s_i)}{\mu(o_i|s_i)}\right)f(s_t,o_t)\right].
\label{eq:app_marginalization}
\end{equation}
Using the factorization
\begin{equation}
\prod_{i=1}^{t}\frac{\pi(o_i|s_i)}{\mu(o_i|s_i)}=\left(\prod_{i=1}^{t-1}\frac{\pi(o_i|s_i)}{\mu(o_i|s_i)}\right)\frac{\pi(o_t|s_t)}{\mu(o_t|s_t)},
\label{eq:app_ratio_split}
\end{equation}
and the deterministic state-transition property of autoregressive LLMs, each state $s_t=(x,o_{<t})$ uniquely identifies its preceding token sequence. Consequently, its state-distribution ratio is
\begin{equation}
\begin{aligned}
\frac{d_t^\pi(s_t|x)}{d_t^\mu(s_t|x)}=\frac{\sum_{\tau_{<t}:s_t}P^\pi(\tau_{<t}|x)}{\sum_{\tau_{<t}:s_t}P^\mu(\tau_{<t}|x)}=\frac{\prod_{i=1}^{t-1}\pi(o_i|s_i)}{\prod_{i=1}^{t-1}\mu(o_i|s_i)}=\prod_{i=1}^{t-1}\frac{\pi(o_i|s_i)}{\mu(o_i|s_i)}.
\end{aligned}
\label{eq:app_state_ratio}
\end{equation}
Therefore,
\begin{equation}
\begin{aligned}
\nabla_\theta\mathcal{J}(\pi)&=\mathbb{E}_{x\sim\mathcal{D}}\left[\sum_{t=1}^{T}\mathbb{E}_{s_t\sim d_t^\mu(\cdot|x),\,o_t\sim\mu(\cdot|s_t)}\left[\frac{d_t^\pi(s_t|x)}{d_t^\mu(s_t|x)}\frac{\pi(o_t|s_t)}{\mu(o_t|s_t)}\nabla_\theta\log\pi(o_t|s_t)A^\pi(s_t,o_t)\right]\right].
\end{aligned}
\label{eq:app_off_policy_gradient}
\end{equation}
This proves Theorem~1.

Moreover, the cumulative importance correction at step $t$ naturally decomposes into a state-distribution correction and an action-level correction:
\begin{equation}
\mathcal{R}_t\triangleq\prod_{i=1}^{t}\frac{\pi(o_i\mid s_i)}{\mu(o_i\mid s_i)}=\underbrace{\frac{d_t^\pi(s_t|x)}{d_t^\mu(s_t|x)}}_{\text{state correction}}\cdot\underbrace{\frac{\pi(o_t\mid s_t)}{\mu(o_t\mid s_t)}}_{\text{action correction}}.
\label{eq:app_trajectory_correction}
\end{equation}

\section{Proof of Theorem~2}

For notational simplicity, we omit the superscript $g$ and consider a fixed
trajectory position $t$. Let
\begin{equation}
r_i=\frac{\pi(o_i\mid s_i)}{\mu(o_i\mid s_i)},\qquad i=1,\ldots,t,
\end{equation}
and suppose that the log-ratios are independent and identically distributed
Gaussian random variables with variance $\sigma^2$. Since $r_i$ is an
importance ratio, its expectation under the behavior policy satisfies
\begin{equation}
\mathbb{E}_{\mu}[r_i]=\mathbb{E}_{o_i\sim\mu(\cdot\mid s_i)}\left[\frac{\pi(o_i\mid s_i)}{\mu(o_i\mid s_i)}\right]=\sum_{o_i}\pi(o_i\mid s_i)=1.
\end{equation}

Let
\begin{equation}
\log r_i\sim\mathcal{N}(m,\sigma^2).
\end{equation}
Using the moment-generating function of a Gaussian random variable,
\begin{equation}
\mathbb{E}_{\mu}[e^{\log r_i}]=e^{m+\sigma^2/2}.
\end{equation}
Since $\mathbb{E}_{\mu}[r_i]=1$, it follows that
\begin{equation}
e^{m+\sigma^2/2}=1,
\end{equation}
and therefore
\begin{equation}
m=-\frac{\sigma^2}{2}.
\end{equation}
Consequently,
\begin{equation}
\log r_i\sim\mathcal{N}\left(-\frac{\sigma^2}{2},\sigma^2\right).
\end{equation}

We first consider the exact trajectory-level correction
\begin{equation}
\mathcal{R}_t=\prod_{i=1}^{t}r_i.
\end{equation}
Taking logarithms transforms the product into a sum:
\begin{equation}
\log\mathcal{R}_t=\sum_{i=1}^{t}\log r_i.
\end{equation}
By the independence assumption, the sum of the Gaussian log-ratios is also
Gaussian, with mean and variance given by the sums of the individual means
and variances:
\begin{equation}
\log\mathcal{R}_t\sim\mathcal{N}\left(-\frac{t\sigma^2}{2},t\sigma^2\right).
\end{equation}

For a Gaussian random variable $X\sim\mathcal{N}(m,v)$, its exponential
$Y=e^X$ is log-normally distributed. Its first and second moments are
\begin{equation}
\mathbb{E}[Y]=\mathbb{E}[e^X]=e^{m+v/2},\quad \mathbb{E}[Y^2]=\mathbb{E}[e^{2X}]=e^{2m+2v}.
\end{equation}
Therefore, its variance is
\begin{equation}
\operatorname{Var}(Y)=\mathbb{E}[Y^2]-\mathbb{E}[Y]^2=e^{2m+2v}-e^{2m+v}=e^{2m+v}(e^v-1).
\end{equation}

Applying this result to $X=\log\mathcal{R}_t$ with
$m=-t\sigma^2/2$ and $v=t\sigma^2$, we obtain
\begin{equation}
\begin{aligned}
\operatorname{Var}_{\mu}(\mathcal{R}_t)=e^{-t\sigma^2+t\sigma^2}(e^{t\sigma^2}-1)
=e^{t\sigma^2}-1.
\end{aligned}
\end{equation}
Thus, the variance of the exact product correction grows exponentially with
the trajectory length $t$.

Next, consider the geometric-mean correction
\begin{equation}
\widetilde{\mathcal{R}}_t=\mathcal{R}_t^{1/t}.
\end{equation}
Its logarithm is
\begin{equation}
\log\widetilde{\mathcal{R}}_t=\frac{1}{t}\log\mathcal{R}_t=\frac{1}{t}\sum_{i=1}^{t}\log r_i.
\end{equation}
Since the log-ratios are independent Gaussian random variables, scaling their
sum by $1/t$ gives
\begin{equation}
\log\widetilde{\mathcal{R}}_t\sim\mathcal{N}\left(-\frac{\sigma^2}{2},\frac{\sigma^2}{t}\right).
\end{equation}
Applying the same log-normal variance formula with
$m=-\sigma^2/2$ and $v=\sigma^2/t$ yields
\begin{equation}
\operatorname{Var}_{\mu}(\widetilde{\mathcal{R}}_t)=e^{-\sigma^2+\sigma^2/t}\left(e^{\sigma^2/t}-1\right).
\end{equation}

To characterize its asymptotic behavior, we use the Taylor expansion
\begin{equation}
e^x-1=x+O(x^2),\qquad x\rightarrow0.
\end{equation}
Setting $x=\sigma^2/t$ gives
\begin{equation}
e^{\sigma^2/t}-1=\frac{\sigma^2}{t}+O\left(\frac{1}{t^2}\right).
\end{equation}
Moreover,
\begin{equation}
e^{-\sigma^2+\sigma^2/t}=e^{-\sigma^2}\left(1+O\left(\frac{1}{t}\right)\right).
\end{equation}
Multiplying the two terms, we obtain
\begin{equation}
\operatorname{Var}_{\mu}(\widetilde{\mathcal{R}}_t)=e^{-\sigma^2}\frac{\sigma^2}{t}+O\left(\frac{1}{t^2}\right).
\end{equation}
Hence,
\begin{equation}
\operatorname{Var}_{\mu}(\widetilde{\mathcal{R}}_t)=O\left(\frac{1}{t}\right).
\end{equation}

In contrast, the exact product correction satisfies
\begin{equation}
\operatorname{Var}_{\mu}(\mathcal{R}_t)=e^{t\sigma^2}-1=\Theta(e^{t\sigma^2}),
\end{equation}
which completes the proof.

\section{Proof of Theorem~3}
\label{app:concentration_correction}

Let $X_i=\log r_i$ and define the centered variables
$Y_i=X_i-\mathbb{E}[X_i]$ for $i=1,\ldots,t$. By assumption, the $Y_i$ are
independent and satisfy
\begin{equation}
\mathbb{E}\left[e^{\lambda Y_i}\right]\leq\exp\left(\frac{\lambda^2\sigma^2}{2}\right),\qquad \forall\lambda\in\mathbb{R}.
\label{eq:app_subgaussian_assumption}
\end{equation}

Consider the centered sample mean
$\overline{Y}_t=\frac{1}{t}\sum_{i=1}^{t}Y_i$. Independence gives
\[
\mathbb{E}\left[e^{\lambda\overline{Y}_t}\right]
=\prod_{i=1}^{t}\mathbb{E}\left[e^{\lambda Y_i/t}\right]
\leq\exp\left(\frac{\lambda^2\sigma^2}{2t}\right),
\]
where the inequality follows by applying Eq.~\eqref{eq:app_subgaussian_assumption}
with $\lambda/t$ to each $Y_i$. Thus, $\overline{Y}_t$ is
sub-Gaussian with variance proxy $\sigma^2/t$.

For any $\lambda>0$, Markov's inequality yields
\[
\Pr\left(\overline{Y}_t\geq\epsilon\right)
=\Pr\left(e^{\lambda\overline{Y}_t}\geq e^{\lambda\epsilon}\right)
\leq e^{-\lambda\epsilon}\mathbb{E}\left[e^{\lambda\overline{Y}_t}\right]
\leq\exp\left(-\lambda\epsilon+\frac{\lambda^2\sigma^2}{2t}\right).
\]
The exponent is minimized at $\lambda^\star=t\epsilon/\sigma^2$, which gives
\begin{equation}
\Pr\left(\overline{Y}_t\geq\epsilon\right)\leq\exp\left(-\frac{t\epsilon^2}{2\sigma^2}\right).
\label{eq:app_upper_tail}
\end{equation}

Since the sub-Gaussian condition holds for all $\lambda\in\mathbb{R}$,
$-Y_i$ is also $\sigma^2$-sub-Gaussian. Applying the same argument to
$-\overline{Y}_t$ gives
\begin{equation}
\Pr\left(\overline{Y}_t\leq-\epsilon\right)\leq\exp\left(-\frac{t\epsilon^2}{2\sigma^2}\right).
\label{eq:app_lower_tail}
\end{equation}

Using the union bound together with Eqs.~\eqref{eq:app_upper_tail} and
\eqref{eq:app_lower_tail}, we obtain
\begin{equation}
\Pr\left(|\overline{Y}_t|\geq\epsilon\right)\leq2\exp\left(-\frac{t\epsilon^2}{2\sigma^2}\right).
\label{eq:app_two_sided_mean_bound}
\end{equation}

Finally, by the definition of the geometric-mean correction,
\[
\log\widetilde{\mathcal{R}}_t=\frac{1}{t}\sum_{i=1}^{t}X_i.
\]
Therefore,
\[
\overline{Y}_t
=\frac{1}{t}\sum_{i=1}^{t}\left(X_i-\mathbb{E}[X_i]\right)
=\log\widetilde{\mathcal{R}}_t-\mathbb{E}\left[\log\widetilde{\mathcal{R}}_t\right].
\]
Substituting this identity into Eq.~\eqref{eq:app_two_sided_mean_bound} yields
\begin{equation}
\Pr\left(\left|\log\widetilde{\mathcal{R}}_t-\mathbb{E}[\log\widetilde{\mathcal{R}}_t]\right|\geq\epsilon\right)\leq2\exp\left(-\frac{t\epsilon^2}{2\sigma^2}\right),
\end{equation}
which completes the proof.

\end{document}